\documentclass{article}

\usepackage[dblblindworkshop,final]{neurips_2026}
\workshoptitle{Grounded User Simulation for Model Evaluation and Training: Diversity, Fidelity, and Validity}
\usepackage[utf8]{inputenc}
\usepackage[T1]{fontenc}
\usepackage{hyperref}
\usepackage{url}
\usepackage{booktabs}
\usepackage{amsfonts}
\usepackage{nicefrac}
\usepackage{microtype}
\usepackage{xcolor}
\usepackage{graphicx}
\usepackage{siunitx}

\title{Prompts versus Rules: Auditing and Controlling Speech Naturalness Behaviors in Voice User Simulators}

\author{%
  Riqiang Wang \quad Elena Khasanova \quad Harsh Saini \quad Lex Konnelly \\
  \textbf{Parsa Kavehzadeh \quad Matthias Lee \quad Mohamed Attia} \\
  Dialpad Inc. \\
  \texttt{\{riqiang.wang, elena.khasanova, hsaini, lex.konnelly,} \\
  \texttt{parsa.kavehzadeh, matthias.lee, mohamed.attia\}@dialpad.com}
}

\begin{document}

\maketitle

\begin{abstract}

As voice agents gain more popularity commercially, the user simulators used to evaluate the deployed agents are also being developed to include more realistic, variable, and diverse \emph{speech naturalness behaviors} --- disfluency, interruption and backchanneling. The quality of the user simulator directly affects the validity of agent evaluation results. However, we find that most studies so far have not examined in detail whether the intended configuration for these behaviors is realized in the simulation. In this study, we audit the realized naturalness behaviors of $\tau$-Voice, our own LLM-based prompting approach across three models, and our rule-based injection algorithm for disfluency, interruption, and backchanneling. We find that prompting for these behaviors is unreliable and produces speech inconsistent with the instructions, placed and distributed less naturally than the instruction implies. In contrast, our rule-based, model-free algorithm produces controllable and diverse naturalness behaviors more aligned with natural speech. Our results suggest that LLMs not purpose-trained for user simulation are not sufficient on their own to represent authentic user behavior, and that linguistically informed deterministic approaches or specialized models are needed to close the gap; auditing and reporting realized naturalness behaviors, rather than configured settings, is what makes that gap visible.

\end{abstract}

\section{Introduction}
\label{sec:intro}
Customer-facing voice conversational agentic systems are being built faster than the evaluation systems for them \citep{agentsurvey,agentsmatter}. 
Evaluating them properly means accounting for the whole working system, including turn-taking, speech recognition (STT) and synthesis (TTS), aside from the usual large language model (LLM) and tool calls evaluation. The most faithful way to do this is to have people call them, but this cannot run on every change. A common approach is to replace the person with a speech-to-speech LLM-powered user simulator, which lets evaluation scale. 

However, the user simulator itself becomes part of the measurement, and voice conversation is where the gap between the configured simulator settings and the realized behavior matters most. Speech is \textbf{disfluent} in a way distinct from written text \citep{clarkfoxtree,tottie2019}. In spoken communication, a human participant plans their reply before the other party has finished speaking \citep{levinson}, resulting in a short gap between turns of \SIrange{110}{130}{\milli\second} \citep{heldner}. Both parties speak at once routinely, and listeners emit short responses, a behavior called
\textbf{backchanneling}, in order to signal attention without taking the turn or decline a turn transfer \citep{yngve,schegloff82,ward}. There are spontaneous \textbf{interruptions} to deny or force a turn transfer. We call these three phenomena --- disfluency, interruption and backchanneling --- \emph{speech naturalness behaviors}, and they are the subject of this study: can we expect the controls for them to produce user simulator speech consistent with the configured setting?

In previous studies, reporting of realized naturalness behaviors on the user simulator side is lacking. Three widely used harnesses report only the
agent's side of that pair: $\tau$-Voice \citep{tauvoice} publishes an ``Agent
Interruption Rate,'' FDB v3 \citep{fdb3} names its one measured disfluency
statistic ``Filler Rate'' and reports it for the agent, and EVA-Bench \citep{evabench}
scores whether the agent interrupted the user rather than the
reverse. The user simulator's own configuration is published in each
case, but the realized behavior it produced is not.

To address this blind spot, we analyze the realized rates for \emph{disfluency}, \emph{interruption}, and \emph{backchanneling} produced by standard LLM-driven approaches (specifically $\tau$-Voice), and compare them against our proposed deterministic, rule-based algorithms. For disfluency, we measure the filler particle rate per 1,000 words, while for interruption and backchanneling, we present the realized rates per qualifying interruption / backchannel opportunity and per agent turn. We show that prompting an LLM to produce disfluency tokens or make real-time turn-taking decisions is insufficient, demonstrating a severe gap in \emph{controllability} between approaches and between what is configured versus what is realized.

It should be noted that controllability is not the same as \emph{realism}. However, controllability leads to behavioral diversity, which serves as a necessary condition for realism and robust agent evaluation.

In summary, our contributions are twofold: (1) we present an empirical audit of prompt-based and LLM-driven user simulator harnesses over live full-duplex telephony transport; and (2) we introduce a rule-based algorithm for realizing speech naturalness behaviors. This approach achieves monotonically controllable interruption and backchannel rates and wider diversity in disfluency, while eliminating model latency and API cost.

Voic\section{Related work}
\label{sec:related}

\paragraph{Task-Oriented Spoken User Simulators.} Traditional user simulators operate primarily in the text modality or rely on static scripts to evaluate multi-turn goal completion. Recent efforts extend simulation into the spoken domain: CHATS \citep{chats} renders pre-written scripts into spoken agent-to-agent turns via unit language models, while TriageSim \citep{triagesim} prompts LLMs to introduce disfluency in emergency triage text turns against an internal agent. Most notably, SpokenUS \citep{spokenus} introduces a spoken user simulator grounded in task-oriented dialogue (TOD) built on the SpokenTOD dataset, incorporating slot disclosures, emotional prosody, and barge-in mechanisms. However, specialized spoken simulators like SpokenUS face substantial reproducibility hurdles: while underlying datasets are publicly released, model weights and execution runtimes are withheld, preventing external researchers from executing live, interactive turn-taking evaluations.

\paragraph{Voice Evaluation Harnesses and Full-Duplex Benchmarks.} Emerging evaluation harnesses attempt to benchmark voice agents in multi-turn full-duplex settings. $\tau$-Voice \citep{tauvoice} operates a reactive simulated user on a fixed 2.0-second tick schedule, invoking dedicated LLM API calls at each tick to evaluate backchannels and interruptions while relying on text prompts for disfluency. However, its prompt set contains direct instruction conflicts --- prescribing filler words in global instructions while directing the model to avoid them in persona guidelines --- and reports no realized disfluency rates. Full-Duplex-Bench (FDB v3) \citep{fdb3} measures filler rates exclusively on the agent side while driving the user with non-reactive human audio recordings; EchoChain \citep{echochain} flags timing deviations using pre-generated audio; IHBench \citep{ihbench} plans interruption points during dataset construction; and EVA-Bench \citep{evabench} and VAmoS \citep{vamos} leave naturalness controls disabled or unmonitored. Crucially, these harnesses evaluate agent-side metrics without auditing or verifying the realized naturalness behaviors produced by the simulated user.

\paragraph{Spoken Foundation Models and Naturalness Control.} Native audio foundation models process discrete acoustic tokens directly to capture turn-taking dynamics \citep{dgslm,moshi,stepaudio}. Frameworks like PersonaPlex \citep{personaplex} and F-Actor \citep{factor} introduce voice conditioning and report realized backchannel frequencies or instructed-versus-generated correlations ($r = 0.25$--$0.54$). However, end-to-end audio models lack explicit tool-use capabilities, operate in self-dialogue, or run half-duplex. Furthermore, controlling naturalness behaviors through LLM prompting inherits the instruction-following limitations reported for explicit numeric constraints \citep{lengthconstraints,complexbench}; we show in Section~\ref{result:disfluency} that prompted disfluencies saturate rapidly, underdeliver on requested rates, concentrate heavily on turn-initial words, and fail to produce self-initiated same-turn repairs.

Our work addresses these gaps by evaluating a real-time cascaded user simulator operating over live telephony transport against deployed commercial agents. Unlike $\tau$-Voice's latency-heavy 2.0-second polling schedule, our simulator evaluates event-driven rule-based controls at mid-turn pause boundaries, eliminating API overhead while yielding monotonic controllability over interruption and backchannel rates. Rather than relying on unverified LLM prompts, our rule-based injection and cadence rules produce realistic dispersion and positional distribution across utterances. Finally, while proprietary systems like SpokenUS cannot be directly executed today due to unreleased model checkpoints, we establish open auditing protocols and consider these specialized spoken models as essential future work baselines for live full-duplex comparison if model weights become available. 

\section{Experimental setup}
\label{sec:system}

\subsection{User simulator design}
\paragraph{Interactive Roles and Black-Box Setup.}

Our voice user simulator is designed to interact with any conversational voice agent via telephony endpoint in a black box setup: both agent and user only exchange the audio. This reflects settings similar to a human conversation, where participants allocate turns among themselves, one at a time, at points they negotiate as they go \citep{sacks}, with no shared orchestrator to assign the right to speak. Concretely, two distinct entities communicate via full-duplex audio:
\begin{itemize}
    \item \textbf{Simulated User:} A cascaded pipeline (STT for transcribing audio, LLM to generate a turn, TTS module to produce audio) assigned a structured task goal and a persona via LLM system prompts. We chose a cascaded pipeline with intermediate text representations and pre-synthesis hooks to allow deterministic injection of disfluencies. Voice activity detector (VAD) events determine the timing of interruptions and backchanneling. The default STT, LLM and TTS models are all from the \texttt{gpt-4o-mini} family. The VAD is Silero VAD. 
    \item \textbf{Target Agent Under Test:} A conversational agent deployed on commercial telephony infrastructure. The agent operates strictly as a black box; the simulator has no access to its internal transcriptions, model weights, prompts, or generation probabilities.
\end{itemize}

\paragraph{Reacting to the Agent: Voice Activity Triggers and Event Handling}
\label{sec:decide}

Rather than polling on fixed time intervals \citep{tauvoice}, the simulator reacts dynamically to VAD events detected on incoming agent audio streams:
\begin{enumerate}
    \item \textbf{Short Pause (\SI{200}{\milli\second}):} Occurs mid-speech while the agent is speaking. This event can serve as a trigger for potential real-time interruption and backchanneling.
    \item \textbf{Turn-End Silence (\SI{500}{\milli\second}):} Indicates potential completion of an agent turn, triggering standard response generation.
\end{enumerate}
Deciding naturalness behaviors strictly at \SI{200}{\milli\second} mid-speech pauses eliminates continuous polling overhead while maintaining full-duplex responsiveness. Standard turn-end sends the truncated conversation history to a small LLM to decide whether to respond to the agent or wait because the agent is still working. We recognize the latency trade-off of using voice-activity detectors for turn handling. Humans answer each other within \SIrange{110}{130}{\milli\second} \citep{heldner}, whereas our system requires at least \SI{500}{\milli\second} plus the latency of the models. This trade-off is outside the scope of this study.

\subsection{Rule-Based Naturalness Controls}
\label{sec:controls}

\begin{figure}[t]
\footnotesize
\begin{verbatim}
rng = Random(seed=utterance_text)  
p = tendency * MAX_PROB_PER_WORD 

for each word boundary:
    if the word is a part of an email, phone number, other alpha numeric string:
        continue  
    if rng.random() < p:
        roll = rng.random()
        if roll < P_FILLER:            insert filler ("um," / "uh,")
        elif roll < P_FILLER + P_REPS: repeat the next word's stem ("I- I need...")
        else:                          insert pause token (not discussed in this paper)
\end{verbatim}
\caption{Rule-based disfluency injection algorithm.}
\label{fig:disfluency}
\end{figure}
\begin{figure}[t]
\footnotesize
\begin{verbatim}
on short_pause:                       # 200 ms of silence, mid-speech
    if interruption_tendency > 0
       and no interruption yet in this agent turn
       and agent_speaking_seconds >= 10.0 - 8.5 * interruption_tendency
       and agent_words >= 12
       and agent_speech is not a question or a request:
        return INTERRUPT              # checked first, and returns immediately

    if backchannel_tendency > 0
       and this is not the agent's first turn
       and agent_speech is not a question or a request
       and agent_words >= (6                                if first in turn
                           else words_at_last_backchannel
                                + 25 - 17 * backchannel_tendency):
        return BACKCHANNEL

    return NOTHING
\end{verbatim}
\caption{Interruption and backchannel decision taken at every short pause. The interruption rule is
evaluated first and returns immediately, so at a pause where it passes the
backchannel rule is never reached --- which is why in our experiment the two behaviors are varied
one at a time. To decide whether the agent's speech is a question or a request, a fixed text pattern match is used on its transcript, e.g. a trailing question mark, utterance starting with "could you...".}
\label{fig:int}
\end{figure}

In this section, we present the rule-based controls for the three naturalness behaviors --- \textbf{disfluency}, \textbf{interruption}, and \textbf{backchanneling} --- which we compare against the prompted instructions. They are governed by rules scaled by configured \emph{tendency} parameters $\alpha, \beta, \gamma \in [0, 1]$. At $0.0$, the behavior is disabled.

\paragraph{Disfluency:} As formalized in Figure~\ref{fig:disfluency}, disfluencies are injected into the generated response text prior to speech synthesis based on tendency $\alpha$. To prevent injected disfluencies from corrupting future conversation context, they are applied strictly as a post-generation text rewrite immediately prior to speech synthesis. Three key design choices govern this algorithm:
\begin{enumerate}
    \item \textbf{Text-Seeded Determinism:} The pseudo-random number generator is seeded directly from the utterance text ($\text{seed} = \text{text}$). Consequently, re-synthesizing or replaying the same LLM response guarantees identical disfluency placement, enabling reproducible benchmark runs.
    \item \textbf{Alphanumeric Protection:} Word boundaries adjacent to alphanumeric words containing digits or the \texttt{@} symbol (e.g., telephone numbers, email addresses) are strictly excluded. Injecting repetitions into these words would artificially degrade the target agent's entity extraction and distort Word Error Rate (WER) references, potentially causing incorrect attribution of error to agent's speech-to-text.
    \item \textbf{Spatial Distribution:} We implement a tunable distribution for different disfluency features: fillers, repetitions / stutters, pause tokens. This allows us to tune preset values based on calibration results. In this study, \verb|MAX_PROB_PER_WORD| is 0.2, \verb|P_FILLER| is 0.5, \verb|P_REPS| is 0.3. We do not discuss the pause token in this study as its realization is dependent on speech synthesis.
\end{enumerate}

\paragraph{Interruption:} Governed by tendency $\beta$, interruption logic evaluates at every $200\text{~ms}$ short pause during agent speech (Figure~\ref{fig:int}). Upon triggering, a short interruption is generated using the agent's truncated transcript as context and immediately synthesized and sent.

\paragraph{Backchanneling:} Governed by tendency $\gamma$, backchannels evaluate at the same $200\text{~ms}$ short pause if the interruption rule does not pass. To eliminate speech synthesis latency, backchannels play immediately from a fixed, pre-synthesized audio inventory of brief listener tokens (e.g., ``mm-hmm'', ``yeah'', ``right'').

\section{Methodology}
\label{sec:method}
\paragraph{Disfluency: Prompted Versus Injected.}
\label{sec:method:disfluency}

Disfluency is the one naturalness behavior measurable without interactive conversational dynamics, enabling evaluation at sample sizes unachievable in full-duplex live studies. We evaluate whether instruction-prompting and rule-based injection effectively control disfluency generation across language models and parameter configurations.

To ensure fair across-condition comparisons, each measured generation evaluates a replayed turn against the deployed conversational agent described in Section~\ref{sec:system}. From 131 successful interactive conversations, we extracted 858 user response positions, sampled 300 positions distributed evenly across dialogue contexts (at most three positions per conversation), and froze this evaluation set. One position is excluded because Gemini's content filter returned an empty candidate for it under all of its conditions; it is dropped from every model and from the injection rows, so all conditions are computed over the same 299 paired positions with identical system contexts and dialogue histories.

We benchmark six prompt conditions per model across three instruction-tuned language models --- \texttt{gpt-4o-mini}, \texttt{gpt-5}, and \texttt{Gemini 3 Pro Preview} --- together with four model-independent injection settings. Experimental conditions include:
\begin{itemize}
    \item \textbf{Uninstructed baseline}: standard text generation without disfluency instructions; 
    \item \textbf{System prompt directives}: system instructions requesting disfluency at three strengths (see Appendix \ref{app:disfluency-prompt}).
    \item \textbf{Published harness directives}: verbatim prompt directives from $\tau$-Voice \citep{tauvoice} (see Appendix \ref{app:disfluency-prompt}).
    \item \textbf{Rule-based injection}: the pre-synthesis text rewrite described in Section~\ref{sec:controls}, at four $\alpha$ values.

\end{itemize}

\paragraph {Interruption and Backchanneling: Empirical Analysis.}
\label{sec:method:interrupt}

Interruption and backchanneling exist strictly within interactive multi-turn dialogue, so we measure their realized rates during live full-duplex runs against the deployed voice agent described in Section~\ref{sec:system}.
In human conversation, interruption/backchannel rates are meaningfully evaluated against an \textbf{opportunity denominator}: a valid mid-turn moment where an interruption/backchannel can genuinely occur. We define opportunity as a $200\text{~ms}$ mid-speech pause in agent audio, excluding legitimate turn-end silences ($\ge 500\text{~ms}$) and periods where the user simulator itself is outputting audio. For the LLM-driven framework in $\tau$-Voice, an opportunity corresponds directly to the number of times the interruption/backchannel LLM-based classifiers are consulted. In $\tau$-Voice, a reactive simulated user operates on a fixed $2.0\text{~s}$ tick schedule (where each tick represents $0.2\text{~s}$ of audio, triggering classifier consultations every 10 ticks). We measure and report interruption/backchannel rates both per opportunity and per agent turn. See Appendix \ref{app:int-prompt} for the verbatim prompts used by $\tau$-Voice.

We contrast the realized behavior of (1) the LLM-driven approach in $\tau$-Voice \citep{tauvoice} with (2) our rule-based controls (Section~\ref{sec:controls}). For (1), we run the benchmark on the Retail dataset and sample 20 successful multi-turn full-duplex sessions, collecting tick-level execution counts. For (2), we evaluate seven paired conditions across 20 sessions each.

\section{Results and Analysis}
\label{sec:result}
\subsection{Disfluency: Prompting versus Rule-Based Injection}
\label{result:disfluency}
Table~\ref{tab:categories} reports the realized rates across all conditions and models.
\paragraph{Rates:} Without explicit disfluency instructions, \textit{baseline} instruction-tuned models produce virtually zero fillers per 1,000 words. We attribute this to instruction tuning and reinforcement learning from human feedback penalizing non-fluent token generation, although we cannot verify this on closed models. With disfluency prompting, filler rates increase monotonically with requested frequency on two of the three models. However, prompting underdelivers relative to the requested scale: stepping from ``occasionally'' to ``roughly half'' produces a large increase in filler rate, which then almost plateaus going from ``roughly half'' to ``most sentences''. In contrast, \textit{injection} scales close to linearly for fillers and slightly faster than linearly for repetitions.
\paragraph{Types:} Prompting generates filler particles almost exclusively, failing to produce repetitions or false starts, with the exception of \texttt{Gemini 3 Pro Preview}. Repetition rates reach only $2.96$ per 1,000 words at maximum prompt strength on \texttt{gpt-4o-mini}, compared to $47.06$ per 1,000 words under rule-based injection ($\alpha = 1.00$). Furthermore, longer restarts (self-initiated repairs) are completely absent from the outputs despite being listed in the instruction prompt (see Appendix~\ref{app:disfluency-prompt}). Note that they were not included in the rule-based injection design.
\paragraph{Position:} \textit{Prompted} outputs heavily front-load the filler particles, which does not reflect natural human speech. Psycholinguistic research demonstrates that natural speech disfluencies are distributed across utterances in response to dynamic cognitive planning demands \citep{clarkfoxtree,shriberg}. While disfluencies frequently appear sentence-initially during initial utterance formulation \citep{shriberg}, filler particles (\texttt{uh} and \texttt{um}) also regularly emerge at mid-clause boundaries and pre-lexical retrieval points when speakers encounter complex syntactic structures or memory search delays \citep{clarkfoxtree}. However, instruction-prompted language models place filler particles almost exclusively on the first word of a turn. Turn-initial fillers account for $62.8\%\text{--}67.3\%$ of all fillers on \texttt{gpt-4o-mini}, $52.6\%\text{--}60.3\%$ on \texttt{Gemini 3 Pro Preview}, and $71.9\%$ under the $\tau$-Voice prompt on \texttt{gpt-4o-mini}. The model satisfies the disfluency instruction immediately at the start of the sentence and generates the remainder of the sentence with uninterrupted fluency (see Table \ref{tab:categories}). \texttt{gpt-5} is the exception on both counts: it front-loads less and places more of its fillers later in the turn. 

In contrast, \textit{injection} provides both spatial flexibility and realistic mid-utterance distribution. Across operational injection settings ($\alpha = 0.25, 0.50, 0.75, 1.00$), turn-initial fillers account for $11.1\%$, $15.6\%$, $14.3\%$, and $13.6\%$ of total fillers, respectively. As the configured injection rate $\alpha$ scales, the turn-initial share initially increases ($11.1\% \to 15.6\%$) and then decreases ($15.6\% \to 14.3\% \to 13.6\%$). This non-monotonic trajectory reflects the two site types scaling differently: the turn-initial slot can fire at most once per turn, while increasing $\alpha$ continues to populate word boundaries throughout the rest of the sentence. As a result, injection distributes disfluencies across clause boundaries and internal lexical sites. Furthermore, because injection operates via explicit probabilistic rules rather than emergent decoding, the turn-initial insertion probability functions as a tunable hyperparameter that can be calibrated directly to match target human dialogue distributions.

\paragraph{Naturalness:} To quantify the unnaturalness of the prompted outputs, we measure the index of dispersion, defined as the variance-to-mean ratio ($V/M = \sigma^2 / \mu$) of per-turn filler counts. A theoretical Poisson baseline representing independent random occurrences yields $V/M = 1.0$. We observe that \textit{prompted} generations exhibit unnatural uniformity across turns, yielding variance-to-mean dispersion ratios of $0.6\text{--}1.0$ across conditions, compared to $1.15\text{--}1.29$ for rule-based \textit{injection} (see Appendix \ref{app:dispersion} for more details). 

\paragraph{Inter-model variance:} Furthermore, identical system prompts yield realized filler rates that differ by a factor of $1.92\times$ to $2.27\times$ across models at a fixed prompt strength, showing that prompt instructions are not portable rate settings. More specifically, the turn-initial filler share for \texttt{gpt-5} is much lower than for the other models, while \texttt{Gemini 3 Pro Preview}'s repetition rate is much higher. It suggests that even though currently the models overall do not perform well through instruction prompting, correct behavior can be learned through fine-tuning with appropriate data.

\begin{table}[t]
  \centering
  \footnotesize
  \setlength{\tabcolsep}{4pt}
  \caption{Disfluency generation metrics across prompt directives and injection ($n=299$ paired dialogue positions per model and condition). Filler and repetition rates are reported per 1,000 words. Turn-initial share represents the percentage of filler particles occurring on the first word of the turn. Pre-TTS injection operates directly on uninstructed model outputs, producing identical text modifications across models.}
  \label{tab:categories}
  \begin{tabular}{lrrrrrrrrr}
    \toprule
    & \multicolumn{3}{c}{Fillers / 1k} & \multicolumn{3}{c}{Repetitions / 1k}
    & \multicolumn{3}{c}{Turn-Initial Fillers (\%)} \\
    \cmidrule(lr){2-4}\cmidrule(lr){5-7}\cmidrule(lr){8-10}
    Condition & 4o-mini & gpt-5 & Gemini & 4o-mini & gpt-5 & Gemini
              & 4o-mini & gpt-5 & Gemini \\
    \midrule
    Uninstructed Baseline & 0.00 & 0.55 & 0.00 & 0.00 & 0.28 & 0.49 & --- & 0.0 & --- \\
    ``occasionally''      & 19.29 & 24.97 & 37.11 & 0.79 & 1.31 & 8.37 & 67.3 & 17.9 & 52.6 \\
    ``roughly half''      & 33.22 & 75.42 & 64.05 & 1.51 & 6.21 & 22.87 & 64.8 & 28.2 & 56.8 \\
    ``most sentences''    & 34.72 & 73.67 & 75.28 & 2.96 & 6.16 & 32.66 & 62.8 & 37.6 & 60.3 \\
    $\tau$-Voice Prompt   & 12.89 & 12.33 & 11.74 & 0.81 & 0.00 & 0.82 & 71.9 & 35.7 & 60.5 \\
    \midrule
    Injection ($\alpha=0.25$) & \multicolumn{3}{c}{28.41} & \multicolumn{3}{c}{9.47} & \multicolumn{3}{c}{11.1} \\
    Injection ($\alpha=0.50$) & \multicolumn{3}{c}{46.71} & \multicolumn{3}{c}{19.91} & \multicolumn{3}{c}{15.6} \\
    Injection ($\alpha=0.75$) & \multicolumn{3}{c}{64.89} & \multicolumn{3}{c}{31.15} & \multicolumn{3}{c}{14.3} \\
    Injection ($\alpha=1.00$) & \multicolumn{3}{c}{83.78} & \multicolumn{3}{c}{47.06} & \multicolumn{3}{c}{13.6} \\
    \bottomrule
  \end{tabular}
\end{table}

\subsection{Interruption and Backchanneling}
\label{sec:result-realized}

Table~\ref{tab:primary} show the full results of the interruption and backchanneling experiment.

\begin{table}[t]
  \centering
  \small
  \caption{Realized interruption and backchannel rates against configured settings across 20 paired scenarios per condition, evaluated against qualifying mid-turn pause opportunities and agent turns. Note that for $\tau$-Voice, Actions Taken also counts the simulator's own interruption and backchannel actions, not agent side cut-offs.}
  \label{tab:primary}
  \begin{tabular}{lrrrrr}
    \toprule
    System Configuration & Opportunities & Actions Taken & \% of Opp. & Agent Turns & Per Turn \\
    \midrule
    \multicolumn{6}{l}{\textbf{Engine Interruption ($\beta$)}} \\
    $\beta = 0.0$ & 230 & 0 & 0.0\% & 118 & 0.000 \\
    $\beta = 0.3$ & 244 & 36 & 14.8\% & 153 & 0.235 \\
    $\beta = 0.7$ & 246 & 44 & 17.9\% & 153 & 0.288 \\
    $\beta = 1.0$ & 194 & 57 & 29.4\% & 106 & 0.538 \\
    \midrule
    \multicolumn{6}{l}{\textbf{Engine Backchannel ($\gamma$)}} \\
    $\gamma = 0.0$ & 182 & 0 & 0.0\% & 118 & 0.000 \\
    $\gamma = 0.3$ & 188 & 78 & 41.5\% & 143 & 0.545 \\
    $\gamma = 0.7$ & 214 & 94 & 43.9\% & 154 & 0.610 \\
    $\gamma = 1.0$ & 187 & 103 & 55.1\% & 122 & 0.844 \\
    \midrule
    \multicolumn{6}{l}{\textbf{$\tau$-Voice Benchmark} (20 conversations, 346 agent turns)} \\
    Interruption & 1,277 & 219 & 17.1\% & 346 & 0.633 \\
    Backchannel & 1,277 & 13 & 1.0\% & 346 & 0.038 \\
    \bottomrule
  \end{tabular}
\end{table}

\paragraph{$\tau$-Voice Interruption Dynamics:} Our analysis reveals an extreme operational imbalance in $\tau$-Voice's interruption mechanics. Across 346 agent turns, $\tau$-Voice executed 219 interruptions, yielding a realized rate of $0.633$ interruptions per agent turn ($11.0$ per conversation). $\tau$-Voice therefore operates with a highly aggressive persona: it produces close to two interruptions for every three agent turns. The target agent is rarely permitted to complete extended speech turns without mid-sentence cut-offs, which affects evaluation fidelity. Conversely, backchanneling in $\tau$-Voice is virtually non-existent. Out of 1,277 classifier consultations (averaging $64.0$ consultations per conversation and $3.69$ per agent turn), the backchannel classifier returned an affirmative decision only 16 times, of which 13 were realized; the other 3 were overridden by an interruption decided in the same tick. This produces a realized backchannel rate of just $0.038$ per agent turn ($0.65$ per conversation), representing an opportunity conversion rate of barely $1.02\%$ ($13 / 1,277$).

We argue that this combination severely degrades simulation fidelity. While the simulated user incessantly interrupts the target agent, it almost never provides standard listener feedback to indicate attentive listening. Moreover, because $\tau$-Voice queries external classifiers on a fixed $2.0\text{~s}$ schedule whenever the agent is speaking, this silent backchannel mechanism incurs continuous API billing overhead ($1,277$ consultations across 20 calls) while delivering virtually no realized behavioral effect.

\paragraph{Rule-Based Naturalness Controls:} Our rule-based controls provide smooth, monotonic scaling across all configured settings. As interruption tendency $\beta$ scales from $0.3$ to $1.0$, the realized interruption rate increases predictably from $0.235$ to $0.538$ per agent turn ($14.8\%$ to $29.4\%$ of available opportunities). Similarly, scaling backchannel tendency $\gamma$ expands the realized rate from $0.545$ to $0.844$ per turn ($41.5\%$ to $55.1\%$ of opportunities). Every setting differs from baseline under a two-sided Wilcoxon signed-rank test on per-scenario counts (all $p < 10^{-3}$ unadjusted; see Appendix~\ref{app:stats}). In contrast to $\tau$-Voice's one-sided behavioral profile, the rule-based controls allow balanced, controllable naturalness behavior. While the monotonic control is guaranteed, we find that the mapping from a configured parameter setting to realized behavior is non-linear. This occurs because deterministic patience and cadence thresholds interact dynamically with the acoustic speech distribution of the target agent (which exhibits unbroken speech with a median of $7.9\text{~s}$).

Parameter scaling changes the potential decision points in time within an utterance rather than simply multiplying event output. At lower interruption settings ($\beta = 0.3$), the patience threshold requires extended agent speech before an interruption can trigger. As a result, interruptions at $\beta = 0.3$ take the first available mid-speech pause in only $13.9\%$ of instances, landing primarily on an agent turn's third or fourth pause. When we increase tendency to $\beta = 1.0$, the patience threshold collapses, causing the simulator to seize the very first mid-turn pause in $56.1\%$ of cases. Scaling the control parameter fundamentally changes \emph{where} the interruption occurs relative to turn onset.

We also analyze the relationship between interruptions and the opportunity denominator. Under our maximum interruption condition ($\beta = 1.0$), total mid-turn pause opportunities decrease to $194$, compared to $230$ at baseline ($\beta = 0.0$) and $246$ at $\beta = 0.7$. This contraction occurs because an executed interruption forcibly terminates the ongoing agent turn, thereby eliminating subsequent mid-speech pauses that would have occurred had the agent continued speaking. However, we consider this shrinking denominator phenomenon to be a weak effect. 

\section{Conclusion}
\label{sec:conclusion}

In this study, we audited the substantial gap between configured naturalness settings and realized speech behaviors in real-time spoken user simulators, and proposed a rule-based approach to naturalness controls. Evaluating commercial voice agents requires user simulators capable of producing realistic, verifiable and reproducible naturalness behaviors such as disfluency, interruptions, and backchannels over live full-duplex audio connections. However, we found that delegating these naturalness controls to LLMs introduces severe operational shortfalls. Prompted disfluencies underdeliver on target rates, saturate rapidly, concentrate unrealistically on turn-initial tokens, and fail to generate self-initiated same-turn repairs. LLM-decided controls display aggressive interruption behavior with near-total suppression of backchannels, while incurring continuous API polling costs.

In contrast, our proposed rule-based injection algorithm restores monotonic controllability, captures realistic mid-utterance spatial distribution, and achieves natural over-dispersed turn-to-turn variance for disfluencies while eliminating latency and extra model invocation costs. Because the mapping from configured parameters to realized naturalness behavior is inherently non-linear and bound to the speech cadence of the speaker, evaluation harnesses must explicitly audit and report realized naturalness behaviors relative to opportunity baselines rather than relying on unverified simulator configurations or agent-side metrics alone.

Moreover, LLM performance varies greatly across models. While currently available LLMs are insufficient for fine-grained control of naturalness behaviors, deterministic rule-based mechanisms provide a robust alternative. In the long term, training LLMs focused on naturalness behaviors using targeted spoken dialogue data remains a promising direction for future work.

\section*{Limitations}

We acknowledge several limitations in our experimental methodology and system design:

\begin{itemize}
    \item \textbf{Univariate Condition Evaluation:} In our realized behavior experiments (Section~\ref{sec:result-realized}), interruption ($\beta$) and backchannel ($\gamma$) parameters were varied one at a time while holding the complementary parameter at zero. Because our event engine evaluates the interruption rule prior to the backchannel rule at each $200\text{~ms}$ pause, an executed interruption forcibly truncates the turn and pre-empts subsequent backchannel opportunities. Further empirical work is required to model the produced rates when both behaviors are simultaneously active.
    \item \textbf{Target Agent Response Time Sensitivity:} Our patience and cadence thresholds were calibrated against the speech distribution of a specific deployed commercial agent (median unbroken speech duration of $7.9\text{~s}$). Because interruption opportunities depend on the acoustic pause profile of the interlocutor, setting calibration cannot be assumed to transfer directly to agents exhibiting substantially faster or slower conversational rhythms.
    \item \textbf{Syntactic Phrase Splitting in Injection:} While disfluency injection avoids positional front-loading and produces realistic statistical dispersion, selecting insertion points strictly from word boundaries can occasionally split tightly bound syntactic units (e.g., placing fillers between a verb and object pronoun). Additionally, rule-based text injection cannot generate complex multi-word self-initiated repairs without explicit phrase-level substitution logic.
    \item \textbf{Acoustic Latency and Turn-Taking Thresholds:} Operating over production telephony infrastructure introduces network transport jitter and processing delays. Our turn-handling thresholds ($200\text{~ms}$ for mid-speech pauses, $500\text{~ms}$ for turn-end detection) remain noticeably higher than human turn-transition gaps ($110\text{--}130\text{~ms}$), representing a real-time constraint inherent in cascaded audio pipelines.
    \item \textbf{No Human Reference And Agnet-Side Impact:} We do not compare realized rates or positions against their distributions in human conversation. We also do not measure how the modified simulator behavior changes the evaluated agent's performance. Future work should verify the behaviors against human conversation data and measure their end-to-end effect on agent evaluation results.
    \item \textbf{Limited Scope:} We evaluate the results obtained from three models, only in cascade user simulator architecture with specific choices, against a fixed agent, using specific prompting technique. It is possible that the observed effects will not hold under significantly different conditions. While our rule-based approach outperformed other solutions under selected conditions in this study, it presents a promising but not complete solution in respect to modeling disfluency, interruption and backchanneling. Future work should verify the findings under more conditions, including with realtime multimodal models, investigate adaptive thresholding mechanisms to dynamically calibrate patience parameters to the acoustic rhythm of unknown target agents, introduce syntactically aware disfluency injections and complex speech repairs. Ultimately, exploring specialized models explicitly trained on these naturalness behaviors may be necessary to achieve fully faithful and variable user simulation.

\end{itemize}

\begin{ack}
This research was conducted at Dialpad, which provided all compute and resources used in the study. All authors were employees of Dialpad at the time of submission and received no third-party funding in support of this work. The authors declare no other competing interests.
\end{ack}

\bibliographystyle{plainnat}
\bibliography{refs}

\appendix
\section{Verbatim Disfluency Prompt}
\label{app:disfluency-prompt}
Our disfluency prompt, appended to the system prompt. The bracketed frequency is the only text that varies across the three
strengths.

\begin{quote}\small\ttfamily\raggedright
Speech style --- you are speaking aloud on a phone call. [Occasionally (at most
once every few sentences) / Regularly (in roughly half of your sentences) /
Frequently (in most sentences)], include natural spoken disfluencies: filled
pauses ("um", "uh"), brief word repetitions ("I- I need", "the- the account"),
false starts, and mid-sentence pauses marked with "..." or "---". Spread them
naturally rather than in every phrase, and never mention or explain them.
\end{quote}

The disfluency prompts of $\tau$-Voice's user-simulator prompt
\citep{tauvoice}, as quoted in their appendix.

\begin{quote}\small\ttfamily\raggedright
Speak naturally, as on a phone call:\\
- Disfluencies: 'um', 'uh', 'you know', 'like', 'I mean'\\
- Restarts: 'Can you [pause] sorry, I meant to ask, can you help me with...'\\
- Use em dashes (---) and [pause] to signify pauses
\end{quote}

From the same appendix, and contradicting the above.

\begin{quote}\small\ttfamily\raggedright
MINIMAL VERBOSITY: Avoid filler words, pleasantries, or elaboration unless
specifically needed.
\end{quote}

\paragraph{The ablation condition.} Because those two lines conflict, a low
as-published rate has two possible causes: the instruction is weak, or the
contradiction cancels it. We therefore ran $\tau$-Voice's prompt twice. The
\emph{as-published} condition includes the MINIMAL VERBOSITY line exactly as their
appendix prints it; the \emph{ablated} condition drops that one line and changes
nothing else. If the rate rose sharply when the line was removed, the shortfall
would be attributable to the contradiction rather than to the instruction. It does
not: the ratio between the two conditions is 1.22 on \texttt{gpt-4o-mini}, 0.94 on
\texttt{gpt-5} and 0.94 on \texttt{Gemini 3 Pro Preview}, so the instruction itself, not the conflicting
line, sets the delivered rate. See Appendix \ref{app:disfluency}.

\section{\texorpdfstring{$\tau$}{τ}-Voice's turn-taking decision prompts}
\label{app:int-prompt}
$\tau$-Voice decides interruption and backchanneling with two separate LLM
classifiers, consulted at the same moment and resolved by priority (interrupt wins).
Their criteria, quoted from
\texttt{src/tau2/user/user\_simulator\_streaming.py}, are what our own thresholds
are compared against. Interruption:

\begin{quote}\small\ttfamily\raggedright
The agent is CURRENTLY speaking [...] should the user interrupt the agent NOW?\\
Consider:\\
- Has the user heard enough to understand what the agent is asking or saying?\\
- Has the user heard enough to have a response, question, or correction ready?\\
- Did the agent just complete the sentence which has all the pertinent
information the user was looking for?\\
- Do NOT repeatedly interrupt the agent if it has spoken only a few words (say
less than 5 words).\\
Respond with ONLY "YES" [...] or "NO" if they should keep listening.
\end{quote}

Backchanneling:

\begin{quote}\small\ttfamily\raggedright
Say YES if:\\
- The agent has completed at least 2 full, substantive sentences in their current
turn (Short phrases like "Thanks for your patience" [...] don't count as
substantive)\\
- The user hasn't spoken or backchanneled recently (check the last 3 exchanges
for ANY user sound including "mm-hmm", "uh-huh", "okay")\\
Say NO if:\\
- The agent just started speaking (fewer than 2 substantive sentences)\\
- The user spoke OR backchanneled within the last 2-3 exchanges\\
- The agent's current turn contains or ends with a question\\
- The agent is wrapping up or about to finish their thought\\
Frequency guidance:\\
- Aim for roughly 1 continuer per 4-6 sentences of extended agent speech\\
- When in doubt, say NO --- silence is also natural
\end{quote}

\section{Disfluency Full Results}
\label{app:disfluency}

\begin{table}[h]
  \centering
  \footnotesize
  \caption{All conditions, all models, $n=299$ paired positions. Rates per 1000
  words. ``Comb.'' is filler particles plus repetitions. Intervals on ``Comb.'' and
  ``Init.'' are 95\% bootstrap over conversations, 2000 resamples, since positions
  from one conversation share a persona and a history. ``Init.'' is the
  fraction of filler particles at the first word of the turn among all fillers; ``Any'' is the fraction
  of turns carrying at least one disfluency. Counting treats an em dash as a word
  separator, since an unspaced dash otherwise fuses a filler to its neighbor, and
  counts three repetition forms: adjacent (``the- the''), filler-separated (``I,
  uh, I''), and two-word phrase (``are you--- are you''). Injection is
  model-independent: it edits recorded replies and makes no call.}
  \label{tab:disfluency-full}
  \begin{tabular}{llrrrrrr}
    \toprule
    Model & Condition & Fill. & Rep. & Comb. & 95\% CI & Init. & Any \\
    \midrule
    \multicolumn{2}{l}{\emph{recorded live run}}
                      &  0.41 &  0.00 &   0.41 & [0.0, 1.3]       & 1.000                &  0.3\% \\
    \midrule
    \texttt{gpt-4o-mini}   & none           &  0.00 &  0.00 &   0.00 & [0.0, 0.0]       & ---                  &  0.0\% \\
                           & occasionally   & 19.29 &  0.79 &  20.08 & [14.3, 26.1]     & 0.673 [0.51, 0.83]   & 16.1\% \\
                           & roughly half   & 33.22 &  1.51 &  34.73 & [28.9, 40.6]     & 0.648 [0.54, 0.75]   & 28.8\% \\
                           & most sentences & 34.72 &  2.96 &  37.68 & [30.2, 45.1]     & 0.628 [0.55, 0.72]   & 29.1\% \\
                           & $\tau$-Voice   & 12.89 &  0.81 &  13.69 & [9.2, 18.5]      & 0.719 [0.56, 0.89]   & 11.4\% \\
                           & \ \ ablated    & 16.73 &  0.00 &  16.73 & [11.8, 22.3]     & 0.643 [0.50, 0.79]   & 13.7\% \\
    \midrule
    \texttt{gpt-5}         & none           &  0.55 &  0.28 &   0.83 & [0.0, 1.9]       & 0.000                &  1.0\% \\
                           & occasionally   & 24.97 &  1.31 &  26.28 & [21.0, 31.9]     & 0.179 [0.10, 0.26]   & 31.4\% \\
                           & roughly half   & 75.42 &  6.21 &  81.63 & [74.5, 89.4]     & 0.282 [0.24, 0.33]   & 78.9\% \\
                           & most sentences & 73.67 &  6.16 &  79.83 & [73.0, 87.8]     & 0.376 [0.32, 0.43]   & 80.6\% \\
                           & $\tau$-Voice   & 12.33 &  0.00 &  12.33 & [7.7, 17.2]      & 0.357 [0.21, 0.51]   & 13.4\% \\
                           & \ \ ablated    & 10.78 &  0.75 &  11.53 & [8.0, 15.6]      & 0.372 [0.21, 0.52]   & 15.1\% \\
    \midrule
    \texttt{Gemini 3 Pro Preview} & none           &  0.00 &  0.49 &   0.49 & [0.0, 1.3]       & ---                  &  0.7\% \\
                           & occasionally   & 37.11 &  8.37 &  45.47 & [41.5, 49.9]     & 0.526 [0.45, 0.60]   & 59.9\% \\
                           & roughly half   & 64.05 & 22.87 &  86.92 & [80.9, 93.7]     & 0.568 [0.51, 0.62]   & 84.9\% \\
                           & most sentences & 75.28 & 32.66 & 107.94 & [100.6, 115.8]   & 0.603 [0.55, 0.66]   & 94.3\% \\
                           & $\tau$-Voice   & 11.74 &  0.82 &  12.55 & [8.9, 16.0]      & 0.605 [0.45, 0.76]   & 15.1\% \\
                           & \ \ ablated    & 11.56 &  0.22 &  11.78 & [8.8, 14.6]      & 0.453 [0.31, 0.58]   & 17.7\% \\
    \midrule
    Injection              & 0.25           & 28.41 &  9.47 &  37.88 & [29.7, 46.2]     & 0.111 [0.04, 0.19]   & 25.1\% \\
                           & 0.50           & 46.71 & 19.91 &  66.62 & [57.5, 75.9]     & 0.156 [0.10, 0.22]   & 41.1\% \\
                           & 0.75           & 64.89 & 31.15 &  96.03 & [85.4, 106.5]    & 0.143 [0.10, 0.20]   & 53.2\% \\
                           & 1.00           & 83.78 & 47.06 & 130.84 & [121.7, 141.3]   & 0.136 [0.10, 0.18]   & 63.9\% \\
    \bottomrule
  \end{tabular}
\end{table}

\section{Dispersion Between Utterances for Disfluency Counts}
\label{app:dispersion}
Dispersion is the variance-to-mean ratio of the per-turn disfluency count, treating
each condition as one speaker. A value of 1 is what independent placement at a
constant rate would give; below 1 means the counts are more even between utterances
than chance, above 1 means they are more clustered. We expect a more clustered distribution to be
more natural in human speech as utterances become disfluent when they are more complex. However, this is not verified with human data. 

\begin{table}[h]
  \centering
  \small
  \caption{Variance-to-mean ratio of the per-turn filler-plus-repetition count,
  $n=299$ turns per row, with a 95\% bootstrap interval over conversations,
  2000 resamples. An interval excluding $1.0$ separates the condition from
  independent placement at a constant rate.}
  \label{tab:dispersion}
  \begin{tabular}{lccc}
    \toprule
    Condition & \texttt{gpt-4o-mini} & \texttt{gpt-5} & \texttt{Gemini 3 Pro Preview} \\
    \midrule
    occasionally    & 0.95 [0.84, 1.07] & 0.79 [0.69, 0.89] & 0.61 [0.53, 0.70] \\
    roughly half    & 0.83 [0.74, 0.92] & 0.64 [0.55, 0.73] & 0.63 [0.53, 0.73] \\
    most sentences  & 1.00 [0.85, 1.15] & 0.60 [0.50, 0.72] & 0.66 [0.58, 0.74] \\
    $\tau$-Voice    & 0.89 [0.85, 0.93] & 0.96 [0.86, 1.09] & 0.89 [0.81, 1.00] \\
    \ \ ablated     & 0.91 [0.83, 1.03] & 0.89 [0.82, 1.01] & 0.86 [0.79, 0.95] \\
    \midrule
    \multicolumn{4}{l}{\emph{Injection (model-independent)}} \\
    0.25 & \multicolumn{3}{c}{1.18 [1.00, 1.36]} \\
    0.50 & \multicolumn{3}{c}{1.15 [0.98, 1.30]} \\
    0.75 & \multicolumn{3}{c}{1.29 [1.07, 1.52]} \\
    1.00 & \multicolumn{3}{c}{1.23 [1.07, 1.41]} \\
    \bottomrule
  \end{tabular}
\end{table}

\section{Statistical Tests}
\label{app:stats}

\paragraph{Interruption and backchanneling.} Each condition is compared against the
shared baseline with a two-sided Wilcoxon signed-rank test on per-scenario counts,
paired by scenario, 20 scenarios per condition. The interval is a 1000-iteration
paired bootstrap on the mean difference. Zero differences are discarded by the test,
so \emph{nonzero} reports how much of the sample it used. $W = 0$ means every nonzero
pair moved in the same direction. Pause events serve as a negative control: no
control targets them, and none of the six conditions moves them.

\begin{table}[h]
  \centering
  \small
  \caption{Paired comparison against baseline, 20 scenarios per condition. $p$ values
  are unadjusted; under Holm correction across the six treatment tests the largest
  becomes $1.8\times10^{-3}$.}
  \label{tab:stats}
  \begin{tabular}{llrrrlrr}
    \toprule
    Condition & Metric & Pairs & Nonzero & Mean $\Delta$ & 95\% CI & $W$ & $p$ \\
    \midrule
    $\beta = 0.3$  & interruptions & 20 & 18 & $+1.80$ & [1.25, 2.45] & 0 & $1.5\times10^{-4}$ \\
    $\beta = 0.7$  & interruptions & 20 & 18 & $+2.20$ & [1.40, 3.05] & 0 & $1.6\times10^{-4}$ \\
    $\beta = 1.0$  & interruptions & 20 & 18 & $+2.85$ & [1.85, 3.85] & 0 & $1.8\times10^{-4}$ \\
    $\gamma = 0.3$ & backchannels  & 20 & 15 & $+3.90$ & [2.30, 5.85] & 0 & $6.4\times10^{-4}$ \\
    $\gamma = 0.7$ & backchannels  & 20 & 14 & $+4.70$ & [2.85, 6.75] & 0 & $9.4\times10^{-4}$ \\
    $\gamma = 1.0$ & backchannels  & 20 & 15 & $+5.15$ & [2.55, 8.15] & 0 & $6.0\times10^{-4}$ \\
    \midrule
    \multicolumn{8}{l}{\emph{negative control}} \\
    all six & pause events & 20 & 17--19 & $-0.4$ to $+5.9$ & spans 0 & 65--89 & $0.24$--$0.93$ \\
    \bottomrule
  \end{tabular}
\end{table}

Comparisons are against baseline only; no test compares adjacent settings, so
monotonicity in Section~\ref{sec:result-realized} is reported as observed rather
than tested.

\paragraph{Disfluency.} Whether the realized rate tracks the dial is tested with
Kendall's $\tau_b$ over per-turn observations, $n = 1196$ paired observations per
comparison (299 positions $\times$ four settings or strengths). $\tau$-Voice's prompt
is excluded, as it is a different instruction rather than a stronger setting of ours.

\begin{table}[h]
  \centering
  \small
  \caption{Trend tests, Kendall's $\tau_b$ between configured strength and realized
  filler-plus-repetition rate.}
  \label{tab:trend}
  \begin{tabular}{lrr}
    \toprule
    Comparison & $\tau_b$ & $p$ \\
    \midrule
    Injection setting $\alpha$        & 0.275 & $3.8\times10^{-31}$ \\
    \texttt{gpt-4o-mini}, clause strength & 0.254 & $1.1\times10^{-23}$ \\
    \texttt{gpt-5}, clause strength       & 0.520 & $2.1\times10^{-108}$ \\
    \texttt{Gemini 3 Pro Preview}, clause strength & 0.599 & $1.1\times10^{-150}$ \\
    \bottomrule
  \end{tabular}
\end{table}

Injection's $\tau_b$ is the lowest of the four despite being monotone by
construction, because ties dominate: the generator is seeded per utterance, so
short turns render identically at adjacent settings and $43\%$ of positions are
identical at all four.

\section{Examples of Disfluency Generation}
\label{app:example}
Drawn from the generations by \texttt{gpt-4o-mini}. ``---'' is the em-dash token which signals a short pause. The realization of the pause depends on the speech synthesis model and is outside the scope of this study. Injection edits the
recorded reply, so its four rows share one source; the prompted rows are separate
generations and differ in content as well as in disfluency. 

\begin{quote}\small
\textbf{\textit{Position} \texttt{nat\_i1/.../8}} \\
\textbf{agent:} The Essentials listing fee is 1\% with a \$4,000 minimum. \\
\textbf{recorded:} I thought the fee was 2.5\%. Are you sure about that? \\
\textbf{none:} I find that hard to believe. I could have sworn it was 2.5\%. Are you sure about that? \\
\textbf{occasionally:} Uh, are you sure about that? I remember it being 2.5\%. \\
\textbf{roughly half:} Um, I think there might be a mistake. I saw it listed as 2.5\% somewhere. Are you sure about that? \\
\textbf{most sentences:} Um, I- I still think it's 2.5\%. Are you sure about that? \\
\textbf{$\tau$-Voice:} Um, are you sure about that? I thought it was 2.5\%. \\[2pt]
\textbf{injection 0.25:} (unchanged from the recorded reply) \\
\textbf{injection 0.50:} (unchanged from the recorded reply) \\
\textbf{injection 0.75:} I thought the fee was 2.5\%. Are you sure uh, about that? \\
\textbf{injection 1.00:} I thought the fee was 2.5\%. Are --- you uh, sure about uh, that?
\end{quote}

\begin{quote}\small
\textbf{\textit{Position} \texttt{nat\_i3/.../13}} \\
\textbf{agent:} Are you still there? \\
\textbf{recorded:} Yes, I'm here. But I'm just a bit uncertain. Are you sure you can't give any advice on whether now is a good time to buy? \\
\textbf{none:} Yes, I am here. But can you confirm that there's no way to know if it's a good time to buy? \\
\textbf{occasionally:} Yes, um, are you sure you can't give any advice? \\
\textbf{roughly half:} Um, yes, I'm just a bit unsure. Are you sure you can't say if it's a good time? \\
\textbf{most sentences:} Yes, I'm here. But um, are you sure you can't give any advice? \\
\textbf{$\tau$-Voice:} Yes, but--- um, are you sure you can't give any advice? \\[2pt]
\textbf{injection 0.25:} Yes, I'm here. But- But I'm just a bit uncertain. Are you sure you can't give any advice on whether now is a good time to buy? \\
\textbf{injection 0.50:} Yes, I'm here. But- But I'm just a bit uncertain. Are you sure you can't give any uh, advice on whether now is a good uh, time to buy? \\
\textbf{injection 0.75:} (as 0.50) \\
\textbf{injection 1.00:} uh, Yes, I'm here. But I'm just a bit uncertain. Are you sure you can't- can't um, give any advice on whether --- now is a good time to buy?
\end{quote}

\section{Ethical Considerations}
\label{app:ethic}
This work introduces a new approach and evaluation practices for auditing voice user simulators to improve the assessment of real-time voice agents over full-duplex telephony transport. 

\paragraph{Positive Impacts.} 
By establishing reliable evaluation practices and new approaches for disfluency, turn-taking, and interruption handling, this work enables developers to build safer, more resilient voice systems for critical applications such as healthcare triage and accessibility services. Furthermore, employing rule-based algorithms reduces the use of LLMs and hence increase the efficiency of user simulation systems.

\paragraph{Negative Impacts \& Mitigations.} 
Because our framework enables precise control over naturalness behaviors (e.g., disfluencies and backchanneling), there is a potential dual-use risk that similar mechanisms could be applied to make automated spam or social engineering robocalls sound more convincingly human, at a lower cost. However, our user simulator is designed specifically as an evaluation system to diagnose agent vulnerabilities. We advocate for responsible deployment and transparency standards, such as mandatory disclosure mechanisms for automated voice systems.

\end{document}